# Causes and Explanations in the Structural-Model Approach: Tractable Cases

**Thomas Eiter**
Institut für Informationssysteme,
Technische Universität Wien
Favoritenstraße 9-11, 1040 Vienna, Austria
eiter@kr.tuwien.ac.at

**Thomas Lukasiewicz***
Dipartimento di Informatica e Sistemistica,
Università di Roma "La Sapienza"
Via Salaria 113, 00198 Rome, Italy
lukasiewicz@dis.uniroma1.it

## Abstract

In this paper, we continue our research on the algorithmic aspects of Halpern and Pearl's causes and explanations in the structural-model approach. To this end, we present new characterizations of weak causes for certain classes of causal models, which show that under suitable restrictions deciding causes and explanations is tractable. To our knowledge, these are the first explicit tractability results for the structural-model approach.

## 1 INTRODUCTION

Dealing with causality is an important issue which emerges in many applications of AI. While this issue has been widely addressed, it is not settled yet, and a number of competing approaches to modeling causality can be found in the literature. Some of them are based on modal non-monotonic logics (developed especially in the context of logic programming), like Geffner's approach [8, 9], which has been inspired by default reasoning from conditional knowledge bases. More specialized modal-logic based formalisms play an important role in dealing with causal knowledge about actions and change; see especially the work by Turner [24] and the references therein for an overview. A different family of approaches evolved from the area of Bayesian networks, such as Pearl's approach to modeling causality by structural equations [1, 6, 20, 21]. In particular, the evaluation of deterministic and probabilistic counterfactuals has been explored [1].

Causality plays an important role in the generation of explanations, which are of crucial importance in areas like planning, diagnosis, natural language processing, and probabilistic inference. Different notions of explanations have been studied quite extensively, see especially [14, 7, 22] for philosophical work, and [19, 23, 15] for work in AI that is related to Bayesian networks. A critical examination of such approaches from the viewpoint of explanations in probabilistic systems is given in [2].

In a recent paper [11], Halpern and Pearl formalized causality using a model-based definition, which allows for a precise modeling of many important causal relationships. Based on a notion of weak causality, they offer appealing definitions of actual causality [12] and of causal explanations [13]. As Halpern and Pearl show, their notions of actual cause and causal explanation, which is very different from the concept of causal explanation in [17, 18, 8], models well many problematic examples in the literature.

The following example from [11, 12, 13] illustrates the structural-model approach. See especially [1, 6, 20, 21, 10] for more details on structural causal models.

**Example 1.1** *(arsonists)* Suppose two arsonists lit matches in different parts of a dry forest, and both cause trees to start burning. Assume now either match by itself suffices to burn down the whole forest. We may model such a scenario in the structural-model framework as follows. We assume two binary background variables $U_1$ and $U_2$, which determine the motivation and the state of mind of the two arsonists, where $U_i$ is 1 iff arsonist $i$ intends to start a fire. We then have three binary variables $A_1$, $A_2$, and $B$, which describe the observable situation, where $A_i$ is 1 iff arsonist $i$ drops the match, and $B$ is 1 iff the whole forest burns down. The causal dependencies between these variables are expressed by functions, which say that the value of $A_i$ is given by the value of $U_i$, and that $B$ is 1 iff either $A_1$ or $A_2$ is 1. These dependencies can be graphically represented as in Fig. 1.

Causes and explanations for events, such as $B = 1$ (the whole forest burns down), are defined by considering the values of variables in the above model and certain hypothetical variants (see Section 2). □

The semantic aspects of causes and explanation in the structural-model approach have been thoroughly studied in [11, 12, 13], while their computational complexity has been analyzed in [3, 5]. As shown there, causes and explanations are complete for the classes $\Sigma_2^P$ and $\Sigma_3^P$ of the Polynomial Hierarchy, and thus intractable in general. As for computation, Hopkins [16] explored search-based strategies for computing actual cases in both the general and restricted settings. However, no tractable cases (apart from trivial instances) were explicitly known so far. In this paper, we fill this gap and make the following major contributions:

- We present a new characterization of weak causes in the structural-model approach, which applies to a class of causal models where the causal dependencies can be hierarchically structured, which we call *decomposable graphs*. Examples of causal models which are covered by this class, considered in Section 5, are causal trees (Section 4) and the more general layered causal graphs (Section 6).

- By exploiting the characterization, we obtain algorithms for deciding weak causes, actual causes, and different notions of explanations as defined for the structural-model approach [11, 13, 5].

- Imposing suitable conditions, the algorithms for deciding weak causes, actual causes etc run in polynomial time. By this way, we obtain several tractability results for the structural-model approach, and in fact, to our knowledge, the first ones which are explicitly derived.

- Furthermore, extending work by Hopkins [16], we discuss how irrelevant variables can be efficiently removed from a causal model when determining weak and actual causes. This can lead to great simplifications, and may speed up the computation considerably.

---
*Alternate address: Institut für Informationssysteme, Technische Universität Wien, Favoritenstraße 9-11, 1040 Vienna, Austria. E-mail: lukasiewicz@kr.tuwien.ac.at.

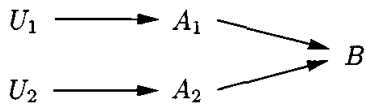

Figure 1: Causal Graph

Note that detailed proofs of all results are given in the extended paper [4].

## 2 PRELIMINARIES

We assume a finite set of *random variables*. Each variable $X_i$ may take on *values* from a finite *domain* $D(X_i)$. A *value* for a set of variables $X = \{X_1, \ldots, X_n\}$ is a mapping $x\colon X \to D(X_1) \cup \cdots \cup D(X_n)$ such that $x(X_i) \in D(X_i)$ (for $X = \emptyset$, the unique value is the empty mapping $\emptyset$). The *domain* of $X$, denoted $D(X)$, is the set of all values for $X$. We say $X$ is *domain-bounded* iff a constant $k$ exists such that $|D(X_i)| \leq k$ for every $X_i \in X$. For $Y \subseteq X$ and $x \in D(X)$, denote by $x|Y$ the restriction of $x$ to $Y$. For disjoint sets of variables $X, Y$ and values $x \in D(X), y \in D(Y)$, denote by $xy$ the union of $x$ and $y$. For (not necessarily disjoint) sets of variables $X, Y$ and values $x \in D(X), y \in D(Y)$, denote by $[x\langle y]$ the union of $x|(X\backslash Y)$ and $y$. We often identify singletons $\{X_i\}$ with $X_i$, and their values $x$ with $x(X_i)$.

### 2.1 CAUSAL MODELS

A *causal model* $M = (U, V, F)$ consists of two disjoint finite sets $U$ and $V$ of *exogenous* and *endogenous* variables, respectively, and a set $F = \{F_X \mid X \in V\}$ of functions $F_X\colon D(PA_X) \to D(X)$ that assign a value of $X$ to each value of the *parents* $PA_X \subseteq U \cup V\backslash\{X\}$ of $X$.

The *causal graph* for $M$, denoted $G(M)$, is the directed graph $(N, E)$, where $N = U \cup V$ and $E = \{(Y, X) \in N \times N \mid Y \in PA_X\}$. Denote by $G_V(M)$ the restriction of $G(M)$ to $V$. A directed graph is *bounded* iff the number of parents of each node is bounded by a constant.

We focus here on the principal class [11] of *recursive* causal models $M = (U, V, F)$ in which a total ordering $\prec$ on $V$ exists such that $Y \in PA_X$ implies $Y \prec X$, for all $X, Y \in V$. In such models, every assignment to the exogenous variables $U = u$ determines a unique value $y$ for every set of endogenous variables $Y \subseteq V$, denoted $Y_M(u)$ (or simply $Y(u)$). In the sequel, $M$ is reserved for denoting a recursive causal model. For any causal model $M = (U, V, F)$, set of variables $X \subseteq V$, and $x \in D(X)$, the causal model $M_x = (U, V, F_x)$, where $F_x = \{F_Y \mid Y \in V\backslash X\} \cup \{F_{X'} = x(X') \mid X' \in X\}$, is a *submodel* of $M$. For $Y \subseteq V$, we abbreviate $Y_{M_x}(u)$ by $Y_x(u)$.

**Example 2.1** *(arsonists continued)* $M = (U, V, F)$ for Example 1.1 is given by $U = \{U_1, U_2\}$, $V = \{A_1, A_2, B\}$, and $F = \{F_{A_1}, F_{A_2}, F_B\}$, where $F_{A_1} = U_1$, $F_{A_2} = U_2$, and $F_B = 1$ iff $A_1 = 1$ or $A_2 = 1$ (Fig. 1 shows the causal graph, i.e., the parent relationships between the variables). □

As for computation, we assume that in $M = (U, V, F)$, every function $F_X\colon D(PA_X) \to D(X)$, $X \in V$, is computable in polynomial time. The following is immediate.

**Proposition 2.1** *For all $X, Y \subseteq V$ and $x \in D(X)$, the values $Y(u)$ and $Y_x(u)$, given $u \in D(U)$, are computable in polynomial time.*

### 2.2 WEAK AND ACTUAL CAUSES

We now recall weak causes from [11, 12]. A *primitive event* is an expression of the form $Y = y$, where $Y$ is a variable and $y$ is a value for $Y$. The set of *events* is the closure of the set of primitive events under the Boolean operators $\neg$ and $\wedge$. The *truth* of an event $\phi$ in $M = (U, V, F)$ under

$u \in D(U)$, denoted $(M, u) \models \phi$, is inductively defined by:

- $(M, u) \models Y = y$ iff $Y_M(u) = y$,
- $(M, u) \models \neg \phi$ iff $(M, u) \models \phi$ does not hold,
- $(M, u) \models \phi \wedge \psi$ iff $(M, u) \models \phi$ and $(M, u) \models \psi$.

We write $\phi(u)$ to abbreviate $(M, u) \models \phi$. For $X \subseteq V$ and $x \in D(X)$, we write $\phi_x(u)$ to abbreviate $(M_x, u) \models \phi$. For $X = \{X_1, \ldots, X_k\} \subseteq V$ with $k \geq 1$ and $x_i \in D(X_i)$, we use $X = x_1 \cdots x_k$ to abbreviate $X_1 = x_1 \wedge \ldots \wedge X_k = x_k$. The following is immediate.

**Proposition 2.2** *Let $X \subseteq V$ and $x \in D(X)$. Given $u \in D(U)$ and an event $\phi$, deciding whether $\phi(u)$ and $\phi_x(u)$ (given $x$) hold can be done in polynomial time.*

Let $M = (U, V, F)$ be a causal model. Let $X \subseteq V$ and $x \in D(X)$, and let $\phi$ be an event. Then, $X = x$ is a *weak cause* of $\phi$ under $u$ iff the following conditions hold:

**AC1.** $X(u) = x$ and $\phi(u)$.

**AC2.** Some set of variables $W \subseteq V \setminus X$ and some values $\bar{x} \in D(X)$, $w \in D(W)$ exist with:

  (a) $\neg \phi_{\bar{x}w}(u)$, and
  
  (b) $\phi_{xw\hat{z}}(u)$ for all $\hat{Z} \subseteq V \setminus (X \cup W)$ and $\hat{z} = \hat{Z}(u)$.

Moreover, $X = x$ is an *actual cause* of $\phi$ under $u$ iff additionally the following minimality condition is satisfied:

**AC3.** $X$ is minimal. That is, no proper subset of $X$ satisfies both AC1 and AC2.

The following result is known.

**Theorem 2.3 (see [3])** *Let $M = (U, V, F)$, $X \subseteq V$, $x \in D(X)$, and $u \in D(U)$. Let $\phi$ be an event. Then, $X = x$ is an actual cause of $\phi$ under $u$ iff $X$ is a singleton and $X = x$ is a weak cause of $\phi$ under $u$.*

**Example 2.2** *(arsonists continued)* Consider the context $u_{1,1} = (1, 1)$ in which both arsonists intend to start a fire. Then, $A_1 = 1$, $A_2 = 1$, and $A_1 = 1 \wedge A_2 = 1$ are weak causes of $B = 1$. In fact, $A_1 = 1$ and $A_2 = 1$ are actual causes of $B = 1$, while $A_1 = 1 \wedge A_2 = 1$ is not. Furthermore, $A_1 = 1$ (resp., $A_2 = 1$) is the only weak cause of $B = 1$ under the context $u_{1,0} = (1, 0)$ (resp., $u_{0,1} = (0, 1)$) in which only arsonist 1 (resp., 2) intends to start a fire. □

### 2.3 EXPLANATION

We now recall the concept of explanation from [11, 13]. Let $M = (U, V, F)$ be a causal model. Let $X \subseteq V$ and $x \in D(X)$, let $\phi$ be an event, and let $C \subseteq D(U)$ be a set of contexts. Then, $X = x$ is an *explanation* of $\phi$ relative to $C$ iff the following conditions hold:

**EX1.** $\phi(u)$ holds, for each context $u \in C$.

**EX2.** $X = x$ is a weak cause of $\phi$ under every $u \in C$ such that $X(u) = x$.

**EX3.** $X$ is minimal. That is, for every $X' \subset X$, some $u \in C$ exists such that $X'(u) = x|X'$ and $X' = x|X'$ is not a weak cause of $\phi$ under $u$.

**EX4.** $X(u) = x$ and $X(u') \neq x$ for some $u, u' \in C$.

**Example 2.3** *(arsonists continued)* Consider the set of contexts $C = \{u_{1,1}, u_{1,0}, u_{0,1}\}$. Then, both $A_1 = 1$ and $A_2 = 1$ are explanations of $B = 1$ relative to $C$, while $A_1 = 1 \wedge A_2 = 1$ is not, as here, the minimality condition EX3 is violated. □

### 2.4 PARTIAL EXPLANATION AND EXPLANATORY POWER

We finally recall the notions of partial and $\alpha$-partial explanation and of explanatory power [11, 13]. Let $M = (U, V, F)$ be a causal model. Let $X \subseteq V$ and $x \in D(X)$, let $\phi$ be an event, and let $C \subseteq D(U)$ be such that $\phi(u)$ holds for all $u \in C$. We use $C^\phi_{X=x}$ to denote the unique largest subset $C'$ of $C$ such that $X = x$ is an explanation of $\phi$ relative to $C'$. The following proposition is easy to see [5].

**Proposition 2.4** *If $X = x$ is an explanation of $\phi$ relative to some $C' \subseteq C$, then $C^\phi_{X=x}$ is defined, and it contains all $u \in C$ such that either $X(u) \neq x$, or $X(u) = x$ and $X = x$ is a weak cause of $\phi$ under $u$.*

Let $P$ be a probability function on $C$, and define

$$P(C^\phi_{X=x} \mid X = x) = \sum_{\substack{u \in C^\phi_{X=x}, \\ X(u) = x}} P(u) \;\Big/\; \sum_{\substack{u \in C, \\ X(u) = x}} P(u).$$

Then, $X = x$ is called an *$\alpha$-partial explanation* of $\phi$ relative to $(C, P)$ iff $C^\phi_{X=x}$ is defined and $P(C^\phi_{X=x} \mid X = x) \geq \alpha$. We say $X = x$ is a *partial explanation* of $\phi$ relative to $(C, P)$ iff $X = x$ is an $\alpha$-partial explanation of $\phi$ relative to $(C, P)$ for some $\alpha > 0$; furthermore, $P(C^\phi_{X=x} \mid X = x)$ is called its *explanatory power* (or *goodness*).

**Example 2.4** *(arsonists continued)* Let $C = \{u_{1,1}, u_{1,0}, u_{0,1}\}$, and let $P$ be the uniform distribution over $C$. Then, both $A_1 = 1$ and $A_2 = 1$ are 1-partial explanations of $B = 1$. That is, both $A_1 = 1$ and $A_2 = 1$ are partial explanations of $B = 1$ with explanatory power 1. □

As for computation, we assume that probability functions $P$ are computable in polynomial time.

## 3 IRRELEVANT VARIABLES

In this section, we describe how an instance of deciding weak cause can be reduced to an equivalent instance in which the (potential) weak cause or the causal model may contain fewer variables. Thus, such reductions remove irrelevant variables in weak causes and causal models.

## 3.1 REDUCING WEAK CAUSES

We first characterize irrelevant variables in weak causes.

The following result shows that deciding whether $X = x$ is a weak cause of $\phi$ under $u$ is reducible to deciding whether $X' = x | X'$ is a weak cause of $\phi$ under $u$, where $X'$ is the set of all $X_i \in X$ that are ancestors of variables in $\phi$.

**Theorem 3.1 (see [5])** *Let $M = (U, V, F)$, $X_0 \in X \subseteq V$, $x \in D(X)$, and $u \in D(U)$. Let $\phi$ be an event. Assume that no directed path in $G(M)$ goes from $X_0$ to a variable in $\phi$, and that $X_0(u) = x(X_0)$. Let $X' = X \setminus \{X_0\}$ and $x' = x | X'$. Then, $X = x$ is a weak cause of $\phi$ under $u$ iff $X' = x'$ is a weak cause of $\phi$ under $u$.*

The next result shows that deciding whether $X = x$ is a weak cause of $\phi$ under $u$ is reducible to deciding whether $X' = x | X'$ is a weak cause of $\phi$ under $u$, where $X'$ is the set of all $X_i \in X$ not "blocked" by some other $X_j \in X$.

**Theorem 3.2** *Let $M = (U, V, F)$, $X_0 \in X \subseteq V$, $x \in D(X)$, and $u \in D(U)$. Let $\phi$ be an event. Assume that every directed path in $G(M)$ from $X_0$ to a variable in $\phi$ contains some $X_i \in X' = X \setminus \{X_0\}$, and that $X_0(u) = x(X_0)$. Let $x' = x | X'$. Then, $X = x$ is a weak cause of $\phi$ under $u$ iff $X' = x'$ is a weak cause of $\phi$ under $u$.*

The following result shows that computing the set of all variables in a weak cause that are not irrelevant according to Theorems 3.1 and 3.2 can be done in linear time.

**Proposition 3.3** *Given $M = (U, V, F)$, $X \subseteq V$, and an event $\phi$,*

*(a) the set $X'$ of all $X_i \in X$ such that $X_i$ is an ancestor in $G(M)$ of a variable in $\phi$ is computable in linear time.*

*(b) the set $X'$ of all variables $X_i \in X$ such that there exists a path from $X_i$ to a variable in $\phi$ that contains no $X_j \in X \setminus \{X_i\}$ is computable in linear time.*

## 3.2 REDUCING CAUSAL MODELS

We next give a characterization of irrelevant variables in causal models, which is essentially due to Hopkins [16].

In the sequel, let $M = (U, V, F)$ be a causal model. Let $X \subseteq V$, $x \in D(X)$, and $u \in D(U)$, and let $\phi$ be an event.

The set of *relevant* variables of $M$ with respect to $X = x$ and $\phi$, denoted $R^\phi_{X=x}(M)$, is the set of all variables $A \in V$ such that either (i), or (ii), or (iii) holds:

(i) $A \in X$, and $A$ is on no directed path in $G(M)$ from a variable in $X \setminus \{A\}$ to a variable in $\phi$.

(ii) $A$ is on a directed path in $G(M)$ from a variable in $X \setminus \{A\}$ to a variable in $\phi$.

(iii) $A$ does not satisfy (i)–(ii), and either $A$ is in $\phi$, or $A$ is a parent of a variable that satisfies (ii).

Note that $X \subseteq R^\phi_{X=x}(M)$. A variable $A \in V$ is *irrelevant* w.r.t. $X = x$ and $\phi$ iff $A \notin R^\phi_{X=x}(M)$. We write $G^\phi_{X=x}(M)$ to denote the restriction of $G(M)$ to $R^\phi_{X=x}(M)$, and often use $G^Y_X(M)$ to abbreviate $G^{Y=y}_{X=x}(M)$.

The reduced causal model of $M$ then does not contain the above irrelevant variables anymore. More formally, the *reduced causal model* of $M = (U, V, F)$ with respect to $X = x$ and $\phi$, denoted $M^\phi_{X=x}$, is the causal model $M' = (U, V', F')$, where $V' = R^\phi_{X=x}(M)$ and

$$F' = \{F'_A = F^\star_A \mid A \in V' \text{ satisfies (i) or (iii)}\} \cup \{F'_A = F_A \mid A \in V' \text{ satisfies (ii)}\},$$

where $F^\star_A$ assigns $A_M(u_A)$ to $A$ for every value $u_A \in D(U_A)$ of the set $U_A$ of all ancestors $B \in U$ of $A$ in $G(M)$.

The following theorem shows that deciding whether $X' = x'$, where $X' \subseteq X$, is a weak cause of $\phi$ under $u$ can be done with respect to $M^\phi_{X=x}$ instead of $M$. This result is a generalization of a similar result by Hopkins [16] for events of the form $X' = x'$ and $\phi = Y = y$, where $X' = X$ and $X', Y$ are singletons.

**Theorem 3.4** *Let $M = (U, V, F)$, $X' \subseteq X \subseteq V$, $x' \in D(X')$, $x \in D(X)$, and $u \in D(U)$, and let $\phi$ be an event. Then, $X' = x'$ is a weak cause of $\phi$ under $u$ in $M$ iff $X' = x'$ is a weak cause of $\phi$ under $u$ in $M^\phi_{X=x}$.*

The following result shows that the reduced causal model and the restriction of its causal graph to the set of endogenous variables can be computed in polynomial and linear time, respectively.

**Proposition 3.5** *Given $M = (U, V, F)$, $X \subseteq V$, $x \in D(X)$, and an event $\phi$, the directed graph $G^\phi_{X=x}(M)$ (resp., causal model $M^\phi_{X=x}$) can be computed in linear (resp., polynomial) time.*

## 4 CAUSAL TREES

In this section, we describe our first class of tractable cases of causes and explanations. More precisely, we show that deciding whether $X = x$ is a weak cause of $Y = y$ under $u$ in $M = (U, V, F)$ is tractable, when $X, Y$ are singletons, $V$ is domain-bounded, and $G^Y_X(M)$ is a bounded directed tree with root $Y$ (see Fig. 2).

Under the same conditions, deciding whether $X = x$ is an actual cause of $Y = y$ under $u$ in $M$, deciding whether $X = x$ is an explanation (resp., a partial explanation or an $\alpha$-partial explanation) of $Y = y$ relative to $\mathcal{C}$ (resp., $(\mathcal{C}, P)$) in $M$, and computing the explanatory power of $X = x$ for $Y = y$ relative to $(\mathcal{C}, P)$ in $M$ are all tractable.

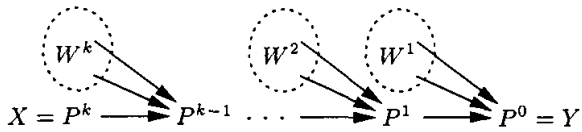

Figure 2: Path from $X$ to $Y$ in a Causal Tree

Observe that the class of tractable cases of causes and explanations described above can be recognized very efficiently. This is shown by the following proposition.

**Proposition 4.1** *Given $M = (U, V, F)$ and $X, Y \in V$, deciding whether $G_X^Y(M)$ is a (bounded) directed tree with root $Y$ can be done in linear time.*

### 4.1 CAUSES

We first focus on deciding weak and actual causes.

In the sequel, let $M = (U, V, F)$ be a causal model, let $X, Y \subseteq V$ be singletons, and let $x \in D(X)$, $y \in D(Y)$, and $u \in D(U)$. Let $G_V(M)$ coincide with $G_X^Y(M)$, and let $G_V(M)$ be a directed tree with root $Y$.

We now give a new characterization of $X = x$ being a weak cause of $Y = y$ under $u$ in $M$, which can be checked in polynomial time under some assumptions. We need some preparation by the following definitions.

Let $X \stackrel{\triangle}{=} P^k \to P^{k-1} \to \cdots \to P^0 \stackrel{\triangle}{=} Y$ be the unique directed path from $X$ to $Y$ in $G_V(M)$. For every $i \in \{1, \ldots, k\}$, denote by $W^i$ the set of all parents of $P^{i-1}$ in $G_V(M)$ that are different from $P^i$ (cf. Fig. 2). For each $i \in \{1, \ldots, k\}$, we define $\hat{p}^i = P^i(u)$.

We define $R^0 = \{D(Y) \setminus \{y\}\}$, and for each $i \in \{1, \ldots, k\}$, we define $R^i$ as follows:

$$R^i = \{ \boldsymbol{p} \subseteq D(P^i) \mid \exists w \in D(W^i) \, \exists \boldsymbol{p}' \in R^{i-1} :$$
$$P_{\hat{p}^i w}^{i-1}(u) \in D(P^{i-1}) \setminus \boldsymbol{p}',$$
$$p \in \boldsymbol{p} \text{ iff } P_{pw}^{i-1}(u) \in \boldsymbol{p}' \}.$$

Intuitively, $R^i$ contains all sets of possible values of $P^i$ in **AC2**(a). Here, $P^0 \stackrel{\triangle}{=} Y$ must be set to a value different from $y$, and the possible values of each other $P^i$ depend on the possible values of $P^{i-1}$. At the same time, the complements of sets in $R^i$ are all sets of possible values of $P^i$ in **AC2**(b). In summary, **AC2**(a) and (b) hold iff some $\boldsymbol{p} \in R^k$ exists such that $\boldsymbol{p} \neq \emptyset$ and $x \notin \boldsymbol{p}$.

This result is more formally expressed by the following theorem, which can be proved by induction on $i \in \{1, \ldots, k\}$.

**Theorem 4.2** *Let $M = (U, V, F)$, $X, Y \in V$, $x \in D(X)$, $y \in D(Y)$, and $u \in D(U)$. Let $G_V(M) = G_X^Y(M)$, let $G_V(M)$ be a directed tree with root $Y$, and let $R^k$ be defined as above. Then, $X = x$ is a weak cause of $Y = y$ under $u$ in $M$ iff ($\alpha$) $X(u) = x$ and $Y(u) = y$, and ($\beta$) some $\boldsymbol{p} \in R^k$ exists such that $\boldsymbol{p} \neq \emptyset$ and $x \notin \boldsymbol{p}$.*

The next theorem shows that deciding whether $X = x$ is a weak cause of $Y = y$ under $u$ in $M$ is tractable, when $X$ and $Y$ are singletons, $V$ is domain-bounded, and $G_X^Y(M)$ is a bounded directed tree with root $Y$. This result follows from Theorem 4.2 and the recursive definition of $R^i$. By Theorem 2.3, the same tractability result holds for actual causes, as the notion of actual cause coincides with the notion of weak cause where $X$ is a singleton.

**Theorem 4.3** *Given $M = (U, V, F)$, $X, Y \in V$, $x \in D(X)$, $y \in D(Y)$, and $u \in D(U)$, where $V$ is domain-bounded, and $G_X^Y(M)$ is a bounded directed tree with root $Y$, deciding whether $X = x$ is a weak (resp., an actual) cause of $Y = y$ under $u$ in $M$ can be done in polynomial time.*

### 4.2 EXPLANATIONS

The following two theorems show that deciding whether $X = x$ is an explanation (resp., a partial explanation or an $\alpha$-partial explanation) of $Y = y$ relative to $\mathcal{C}$ (resp., $(\mathcal{C}, P)$) in $M$, and computing the explanatory power of $X = x$ for $Y = y$ relative to $(\mathcal{C}, P)$ in $M$ are all tractable under the conditions of the previous subsection. These results follow from Proposition 2.2 and Theorem 4.3.

**Theorem 4.4** *Given $M = (U, V, F)$, $X, Y \in V$, $x \in D(X)$, $y \in D(Y)$, and $\mathcal{C} \subseteq D(U)$, where $V$ is domain-bounded, and $G_X^Y(M)$ is a bounded directed tree with root $Y$, deciding whether $X = x$ is an explanation of $Y = y$ relative to $\mathcal{C}$ in $M$ can be done in polynomial time.*

**Theorem 4.5** *Let $M = (U, V, F)$, $X, Y \in V$, $x \in D(X)$, $y \in D(Y)$, $\mathcal{C} \subseteq D(U)$, and $P$ be a probability function on $\mathcal{C}$, such that $V$ is domain-bounded, $G_X^Y(M)$ is a bounded directed tree with root $Y$, and $Y(u) = y$ for all $u \in \mathcal{C}$. Then,*

*(a) deciding if $X = x$ is a partial explanation of $Y = y$ relative to $(\mathcal{C}, P)$ in $M$ is possible in polynomial time.*

*(b) deciding whether $X = x$ is an $\alpha$-partial explanation of $Y = y$ relative to $(\mathcal{C}, P)$ in $M$, for some given $\alpha \geq 0$, can be done in polynomial time.*

*(c) given $X = x$ is a partial explanation of $Y = y$ relative to $(\mathcal{C}, P)$ in $M$, the explanatory power of $X = x$ is computable in polynomial time.*

## 5 DECOMPOSABLE CAUSAL GRAPHS

In this section, we show that the technique of decomposing causal trees for deciding causes and explanations and for computing the explanatory power described in the previous section can be extended to general causal graphs.

Intuitively, the main idea is to decompose the directed graph $G_V(M)$ into a chain of subgraphs along which we can propagate sets of possible values of variables back to the variables in a potential weak cause (see Fig. 3).

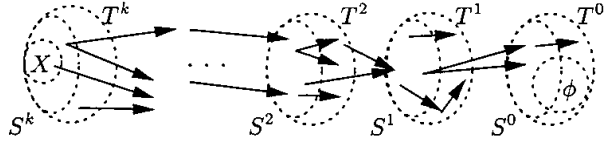

Figure 3: Decomposable Causal Graph

### 5.1 CAUSES

We first concentrate on deciding weak and actual causes.

In the sequel, let $M = (U, V, F)$ be a causal model, let $X \subseteq V$, $x \in D(X)$, and $u \in D(U)$, and let $\phi$ be an event.

Intuitively, to decide whether $X = x$ is a weak cause of $\phi$ under $u$ in $M$, we decompose $G_V(M)$ into a chain of directed subgraphs over the components of an ordered partition $(T^0, \ldots, T^k)$ of $V$, which are connected to each other exactly through some sets $S^0 \subseteq T^0, \ldots, S^k \subseteq T^k$, where every variable in $\phi$ (resp., $X$) belongs to $T^0$ (resp., $S^k$). We then propagate sets of possible values of the $S^i$ in **AC2**(a) and (b) along the chain from $S^0$ to $S^k$. Such a propagation works if certain conditions hold, which are specified in the following concept of a decomposition of $G_V(M)$.

A *decomposition* of $G_V(M)$ with respect to $X = x$ and $\phi$ is a list $((T^0, S^0), \ldots, (T^k, S^k))$ of pairs $(T^i, S^i)$ of sets of endogenous variables such that (D1)–(D6) hold:

**D1.** $(T^0, \ldots, T^k)$ is an ordered partition of $V$.

**D2.** $T^0 \supseteq S^0, \ldots, T^k \supseteq S^k$.

**D3.** Every $A \in V$ occurring in $\phi$ belongs to $T^0$, and $S^k \supseteq X$.

**D4.** For every $i \in \{0, \ldots, k-1\}$, no two variables $A \in T^0 \cup \cdots \cup T^{i-1} \cup T^i \setminus S^i$ and $B \in T^{i+1} \cup \cdots \cup T^k$ are connected by an arrow in $G_V(M)$.

**D5.** For every $i \in \{1, \ldots, k\}$, every child of a variable in $S^i$ in $G_V(M)$ belongs to $(T^i \setminus S^i) \cup S^{i-1}$. Every child of a variable in $S^0$ belongs to $(T^0 \setminus S^0)$.

**D6.** For every $i \in \{0, \ldots, k-1\}$, every parent of a variable in $S^i$ in $G_V(M)$ belongs to $T^{i+1}$. There are no parents of any variable $A \in S^k$.

Such a decomposition is *width-bounded* iff a constant $l$ exists such that $|T^i| \leq l$ for every $i \in \{1, \ldots, k\}$.

Observe that every $M^\phi_{X=x} = (U, V', F')$, where no $A \in X$ is on a path from a variable in $X \setminus \{A\}$ to a variable in $\phi$, has always the trivial decomposition $((V', X))$.

We next define the relations $R^i$, which contain triples $(\boldsymbol{p}, \boldsymbol{q}, F)$, where $\boldsymbol{p}$ (resp., $\boldsymbol{q}$) specifies a set of possible values of $F \subseteq S^i$ in **AC2**(a) (resp., **AC2**(b)).

In detail, we define $R^0$ as follows:

$$R^0 = \{(\boldsymbol{p}, \boldsymbol{q}, F) \mid F \subseteq S^0,\ \boldsymbol{p}, \boldsymbol{q} \subseteq D(F),$$
$$\exists W \subseteq T^0,\ W \cap S^0 = S^0 \setminus F,$$
$$\exists w \in D(W):$$
$$p \in \boldsymbol{p} \text{ iff } \neg \phi_{pw}(u),$$
$$q \in \boldsymbol{q} \text{ iff } \phi_{[q\langle \hat{Z}(u)]w}(u)$$
$$\text{for all } \hat{Z} \subseteq T^0 \setminus (S^k \cup W)\}.$$

For every $i \in \{1, \ldots, k\}$, we then define $R^i$ as follows:

$$R^i = \{(\boldsymbol{p}, \boldsymbol{q}, F) \mid F \subseteq S^i,\ \boldsymbol{p}, \boldsymbol{q} \subseteq D(F),$$
$$\exists W \subseteq T^i,\ W \cap S^i = S^i \setminus F,$$
$$\exists w \in D(W)\ \exists (\boldsymbol{p}', \boldsymbol{q}', F') \in R^{i-1}:$$
$$p \in \boldsymbol{p} \text{ iff } F'_{pw}(u) \in \boldsymbol{p}',$$
$$q \in \boldsymbol{q} \text{ iff } F'_{[q\langle \hat{Z}(u)]w}(u) \in \boldsymbol{q}'$$
$$\text{for all } \hat{Z} \subseteq T^i \setminus (S^k \cup W)\}.$$

We are now ready to give a new characterization of weak cause, which is based on the above concept of a decomposition of $G_V(M)$ and the relations $R^i$.

**Theorem 5.1** *Let $M = (U, V, F)$, $X \subseteq V$, $x \in D(X)$, and $u \in D(U)$. Let $\phi$ be an event. Let $((T^0, S^0), \ldots, (T^k, S^k))$ be a decomposition of $G_V(M)$ with respect to $X = x$ and $\phi$. Let $R^k$ be defined as above. Then, $X = x$ is a weak cause of $\phi$ under $u$ in $M$ iff $(\alpha)$ $X(u) = x$ and $\phi(u)$ holds, and $(\beta)$ some $(\boldsymbol{p}, \boldsymbol{q}, X) \in R^k$ exists such that $\boldsymbol{p} \neq \emptyset$ and $x \in \boldsymbol{q}$.*

The next result shows that deciding whether $X = x$ is a weak (resp., an actual) cause of $\phi$ under $u$ in $M$ is tractable, when $V$ is domain-bounded, and when $G^\phi_{X=x}(M)$ has a width-bounded decomposition provided in the input. This result follows from Theorems 2.3, 3.4, and 5.1 and the recursive definition of the $R^i$'s above.

**Theorem 5.2** *Given $M = (U, V, F)$, $X \subseteq V$, $x \in D(X)$, $u \in D(U)$, an event $\phi$, and a width-bounded decomposition $((T^0, S^0), \ldots, (T^k, S^k))$ of $G^\phi_{X=x}(M)$ with respect to $X = x$ and $\phi$, where $V$ is domain-bounded, deciding whether $X = x$ is a weak (resp., an actual) cause of $\phi$ under $u$ in $M$ can be done in polynomial time.*

### 5.2 EXPLANATIONS

The following two theorems show that deciding whether $X = x$ is an explanation (resp., a partial explanation or an $\alpha$-partial explanation) of $\phi$ relative to $\mathcal{C}$ (resp., $(\mathcal{C}, P)$) in $M$, and computing the explanatory power of $X = x$ for $\phi$ relative to $(\mathcal{C}, P)$ in $M$ are all tractable, when we assume the same restrictions as in Theorem 5.2. These results follow from Proposition 2.2 and Theorem 5.2.

**Theorem 5.3** *Given* $M = (U, V, F)$, $X \subseteq V$, $x \in D(X)$, $\mathcal{C} \subseteq D(U)$, *an event* $\phi$, *and a width-bounded decomposition* $((T^0, S^0), \ldots, (T^k, S^k))$ *of* $G^\phi_{X=x}(M)$ *with respect to* $X = x$ *and* $\phi$, *where* $V$ *is domain-bounded, deciding whether* $X = x$ *is an explanation of* $\phi$ *relative to* $\mathcal{C}$ *in* $M$ *can be done in polynomial time.*

**Theorem 5.4** *Given* $M = (U, V, F)$, $X \subseteq V$, $x \in D(X)$, $\mathcal{C} \subseteq D(U)$, *an event* $\phi$, *a probability function* $P$ *on* $\mathcal{C}$, *and a width-bounded decomposition* $((T^0, S^0), \ldots, (T^k, S^k))$ *of* $G^\phi_{X=x}(M)$ *with respect to* $X = x$ *and* $\phi$, *where* $V$ *is domain-bounded, and* $\phi(u)$ *for all* $u \in \mathcal{C}$,

(a) *deciding if* $X = x$ *is a partial explanation of* $\phi$ *relative to* $(\mathcal{C}, P)$ *in* $M$ *can be done in polynomial time.*

(b) *deciding whether* $X = x$ *is an* $\alpha$-*partial explanation of* $\phi$ *relative to* $(\mathcal{C}, P)$ *in* $M$, *for some given* $\alpha \geq 0$, *can be done in polynomial time.*

(c) *given* $X = x$ *is a partial explanation of* $\phi$ *relative to* $(\mathcal{C}, P)$ *in* $M$, *computing the explanatory power of* $X = x$ *can be done in polynomial time.*

## 6 LAYERED CAUSAL GRAPHS

In general, it is not clear whether causal graphs with width-bounded decompositions can be efficiently recognized, and whether such decompositions can be efficiently computed. In this section, we discuss a large class of causal graphs, called layered causal graphs, that have natural nontrivial decompositions that can be computed in linear time.

Intuitively, such causal graphs $G_V(M)$ can be partitioned into layers $S^0, \ldots, S^k$ such that every arrow goes from a variable in some layer $S^i$ to one in $S^{i-1}$ (see Fig. 4).

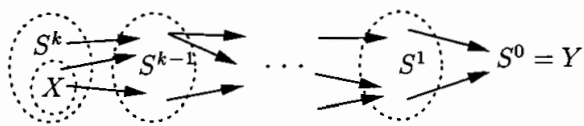

Figure 4: Path from $X$ to $Y$ in a Layered Causal Graph

More formally, let $M = (U, V, F)$ be a causal model, and let $X \subseteq V$, $Y = \{Y_0\} \subseteq V$, $x \in D(X)$, $y \in D(Y)$, and $u \in D(U)$. Then, $G_V(M)$ is *layered* w.r.t. $X$ and $Y$ iff an ordered partition $(S^0, \ldots, S^k)$ of $V$ exists with (L1) and (L2):

**L1.** For every arrow $A \to B$ in $G_V(M)$, there exists some $i \in \{1, \ldots, k\}$ such that $A \in S^i$ and $B \in S^{i-1}$.

**L2.** $Y = S^0$ and $S^k \supseteq X$.

A layered $G_V(M)$ is *width-bounded* for an integer $l \geq 0$ iff there is an ordered partition $(S^0, \ldots, S^k)$ of $V$ with (L1) and (L2) such that $|S^i| \leq l$ for every $i \in \{1, \ldots, k\}$.

The following proposition shows that layered causal graphs $G_V(M)$ have a natural nontrivial decomposition.

**Proposition 6.1** *Let* $M = (U, V, F)$, $X \subseteq V$, $Y = \{Y_0\} \subseteq V$, $x \in D(X)$, *and* $y \in D(Y)$. *Let* $(S^0, \ldots, S^k)$ *be an ordered partition of* $V$ *satisfying (L1) and (L2). Then,* $((S^0, S^0), \ldots, (S^k, S^k))$ *is a decomposition of* $G_V(M)$ *with respect to* $X = x$ *and* $Y = y$.

The next result shows that recognizing layered and width-bounded causal graphs $G_X^Y(M)$ and computing their natural decomposition can be done in linear time.

**Proposition 6.2** *Given* $M = (U, V, F)$, $X \subseteq V$, $Y = \{Y_0\} \subseteq V$, $x \in D(X)$, *and* $y \in D(Y)$, *deciding whether* $G_X^Y(M) = (V', E')$ *is layered and width-bounded for an integer* $l \geq 0$, *and computing the ordered partition* $(S^0, \ldots, S^k)$ *of* $V'$ *with (L1) and (L2) can be done in linear time.*

**Proof (sketch).** Observe that if there is a directed path from a node in $X$ to $Y$, then the ordered partition $(S^0, \ldots, S^k)$ of $V'$ with (L1) and (L2) is unique, if it exists. We can then compute $(S^0, \ldots, S^k) = (T^{-k}, \ldots, T^0)$ as follows. We first compute $G_X^Y(M)$. We then set $\Delta = X$ and $i = 0$. Then, (a) set $T^i$ to the union of $\Delta$ and the set of all parents of a child of $\Delta$, (b) set $\Delta$ to the set of all children of $\Delta$, and (c) decrement $i$. We now repeat (a)–(c) until $\Delta = \emptyset$. Then, $G_X^Y(M)$ is layered iff the computed $T^i$'s are pairwise disjoint, and $G_X^Y(M)$ is width-bounded iff every $|T^i|$ is bounded. This proves the stated result. $\square$

By Proposition 6.1, all the results of Sections 5.1 and 5.2 on causes and explanations in decomposable causal graphs also apply to layered causal graphs as a special case.

It is easy to verify that the relations $R^i$ of Section 5.1 can be simplified as follows for layered causal graphs: We have $R^0 = \{(D(Y) \backslash \{y\}, \{y\}, Y)\}$, and for each $i \in \{1, \ldots, k\}$, the relation $R^i$ is now given as follows:

$$R^i = \{(\boldsymbol{p}, \boldsymbol{q}, F) \mid F \subseteq S^i,\ \boldsymbol{p}, \boldsymbol{q} \subseteq D(F),$$
$$\exists w \in D(S^i \backslash F)\ \exists (\boldsymbol{p}', \boldsymbol{q}', F') \in R^{i-1}:$$
$$p \in \boldsymbol{p} \text{ iff } F'_{pw}(u) \in \boldsymbol{p}',$$
$$q \in \boldsymbol{q} \text{ iff } F'_{[q\langle \hat{Z}(u)]w}(u) \in \boldsymbol{q}' \text{ for all } \hat{Z} \subseteq F \backslash S^k\}.$$

The following theorem is then an immediate corollary of Theorem 5.1 and Proposition 6.1.

**Theorem 6.3** *Let* $M = (U, V, F)$, $X \subseteq V$, $Y \in V$, $x \in D(X)$, $y \in D(Y)$, *and* $u \in D(U)$. *Let* $G_V(M)$ *be layered with respect to* $X$ *and* $Y$, *and let* $R^k$ *be defined as above.*

*Then, $X = x$ is a weak cause of $Y = y$ under $u$ in $M$ iff ($\alpha$) $X(u) = x$ and $Y(u) = y$, and ($\beta$) some $(\mathbf{p}, \mathbf{q}, X) \in R^k$ exists such that $\mathbf{p} \neq \emptyset$ and $x \in \mathbf{q}$.*

The next theorem shows that deciding whether $X = x$ is a weak (resp., an actual) cause of $Y = y$ under $u$ in $M$ is tractable, when $V$ is domain-bounded, and $G_X^Y(M)$ is layered and width-bounded. This result is an immediate corollary of Theorem 5.2 and Proposition 6.1.

**Theorem 6.4** *Let $M = (U, V, F)$, $X \subseteq V$, $Y \in V$, $x \in D(X)$, $y \in D(Y)$, and $u \in D(U)$. If $V$ is domain-bounded, and $G_X^Y(M)$ is layered and width-bounded for a constant $l \geq 0$, then deciding whether $X = x$ is a weak (resp., an actual) cause of $Y = y$ under $u$ in $M$ is possible in polynomial time.*

Similarly, deciding whether $X = x$ is an explanation (resp., a partial explanation or an $\alpha$-partial explanation) of $Y = y$ relative to $\mathcal{C}$ (resp., $(\mathcal{C}, P)$) in $M$, and computing the explanatory power of $X = x$ for $Y = y$ relative to $(\mathcal{C}, P)$ in $M$ are all tractable under the same restrictions. This is immediate by Theorems 5.3 and 5.4 and Proposition 6.1.

## 7 SUMMARY AND OUTLOOK

In this paper, we presented new characterizations of weak causes for certain classes of decomposable causal models, in particular, for causal trees and the more general class of layered causal graphs. By means of these characterizations, we then showed that under suitable restrictions deciding causes and explanations is tractable for these classes. To our knowledge, these are the first explicit tractability results for the structural-model approach. Furthermore, we have also discussed how irrelevant variables can be efficiently removed when deciding causes and explanations.

In this paper, we focused on the problems of deciding causes and explanations. Another important problem is to compute some (resp., all) causes and explanations $X' = x'$ such that $X'$ is contained in a given set of endogenous variables $X$ (cf. [5]). It is not difficult to see that by means of the characterizations that we have obtained, also this computation can be accomplished in polynomial time [4].

An interesting topic of further studies is to explore how to efficiently compute decompositions of causal graphs, and in particular whether there are other important classes of causal graphs different from causal trees and layered causal graphs in which width-bounded decompositions can be recognized and computed efficiently.

### Acknowledgments

This work has been partially supported by the Austrian Science Fund Project Z29-INF and by a Marie Curie Individual Fellowship of the European Community (Disclaimer: The authors are solely responsible for information communicated and the European Commission is not responsible for any views or results expressed).